# Does human-robot trust need reciprocity?*

Joshua Zonca, Alessandra Sciutti

*Abstract*— Trust is one of the hallmarks of human-human and human-robot interaction. Extensive evidence has shown that trust among humans requires reciprocity. Conversely, research in human-robot interaction (HRI) has mostly relied on a unidirectional view of trust that focuses on robots' reliability and performance. The current paper argues that reciprocity may also play a key role in the emergence of mutual trust and successful collaboration between humans and robots. We will gather and discuss works that reveal a reciprocal dimension in human-robot trust, paving the way to a bidirectional and dynamic view of trust in HRI.

## I. Introduction

Humans are inherently social and cooperative beings. This aspect of human behavior is somewhat puzzling, since natural selection should theoretically favor selfish behavior. A crucial mechanism sustaining the emergence and maintenance of cooperation among human is reciprocity, which assumes that one's tendency to cooperate is conditional upon others' cooperation. Reciprocity is also fundamental for the maintenance of mutual trust between peers: if we never trust others, it is unlikely that others will trust us in the future. Indeed, trust among humans is a *relational* phenomenon, which requires that all the individuals involved in interactions and relationships accept a condition of vulnerability to others, believing that others will not exploit this vulnerability [1]. For these reasons, reciprocity has been established in human societies as a social norm [2].

However, reciprocity has not been given a crucial role in human-robot interaction (HRI) research, especially in the study of human-robot trust. The unspoken assumption, which stems from the traditional view of trust in automation, is that the emergence of trust between humans and robots does not need reciprocity due to the intrinsic asymmetrical nature of human-machine relationships. In fact, the research emphasis is almost entirely on the physical, behavioral and functional characteristics of robots: humans trust robots if they are functionally reliable, whereas they do not trust them otherwise.

The main thesis of the current paper is that reciprocity may play a role in supporting mutual trust between humans and robots. In other words, we argue that human trust towards robots may be influenced by the trust expressed, in turn, by robots towards humans during interaction. We will gather existent studies revealing the emergence of reciprocal dynamics in human-robot trust-based interactions and outline a research agenda that see reciprocity as one of the factors shaping human-robot trust and collaboration.

## II. Trust and reciprocity in HRI

Trust is undoubtedly one of the main mechanisms supporting collaboration with robots. Trusting our autonomous partners is crucial to delegate responsibility and accept help from them. Historically, research on trust in HRI conceptualized trust as a one-sided process of evaluation of the functional competence and reliability of robotic agents. Extensive evidence has shown that the main determinant of trust in robots is their performance (e.g. [3, 4]). Humans trust robots as long as they show reliable behavior, but they quickly lose trust in in presence of failures [5, 6], leading to disuse of the robotic system [7, 8].

Nonetheless, evidence in HRI highlighted the emergence of distortions in the process of weighting of robots' competence and the relative expression of trust in them. For instance, recent studies have shown that individuals may over-comply with the instructions of robots, even if they have previously shown faulty or unreliable behavior [9-12]. One possibility is that the overt expression of trust towards robots does not always match the individual internal representation of the robot's reliability. This effect might be driven by pro-social attitudes towards social robots, which have been observed in numerous studies (e.g., see [13-15]). At the same time, recent evidence highlighted the emergence of reciprocity in repeated and multi-stage games such as the Prisoner's Dilemma and the Ultimatum Game [16]. Altogether, we hypothesize that trust-based relationships between humans and robots could be shaped, at least in part, by those relational and reciprocal mechanisms typically intervening in human-human interaction.

A recent study by Zonca and colleagues [17] tested this hypothesis by a novel experimental paradigm investigating the emergence of reciprocal trust in human-robot interaction. In a joint task, a human participant and a humanoid robot iCub made perceptual judgments and signaled their trust in the partner. The robot's trust was dynamically manipulated along the experiment and participants could observe both robots' perceptual responses (that were extremely accurate) and trust feedback. Results show that participants did not learn from a robot that was showing high trust in them, since the robot's trust signaled incompetence. However, they were unwilling to disclose their distrust to the robot if they expected future interactions with it. These findings reveal that the overt expression of trust in robots may be modulated by reciprocity, mirroring recent findings observed in human peer interaction and child-adult interaction [18-21].

*Research supported by the European research Council (ERC Starting Grant 804388, wHiSPER).

J. Z. is with the Italian Institute of Technology, Cognitive Architecture for Collaborative Technologies (CONTACT) Unit, Genoa 16152 Italy (corresponding author, e-mail: joshua.zonca@iit.it).

A. S. is with the Italian Institute of Technology, Cognitive Architecture for Collaborative Technologies (CONTACT) Unit, Genoa 16152 Italy (e-mail: alessandra.sciutti@iit.it).

Strohkorb Sebo and colleagues [22] tested the impact of a robot showing vulnerability on human groups of participants during a collaborative game. Results suggest that robots' vulnerability had a "ripple effect" on the trust-related behavior of participants, who were in turn more willing to disclose their vulnerable state to their teammates, reducing the amount of tension of the team. In line with these findings, a recent study [23] revealed that individuals are more prone to trust a robot and collaborate with it when the robot blames itself for collaborative failures.

In line with this "reciprocal" conceptualization of trust, recent works started to model trust from a robot-centered perspective. In particular, the cognitive architecture developed by Vinanzi and colleagues [24, 25] combines trust and Theory of Mind (TOM) modules with an episodic memory system to allow a humanoid robot to evaluate the trustworthiness of human partners in joint tasks. The authors have shown that allowing robots to monitor the current performance of the human partner(s) and take control of the task in case of need enhances collaborative performance in a joint task.

### III. TOWARDS A "RECIPROCAL" VIEW OF TRUST IN HRI

Altogether, recent research in HRI put the accent on a bidirectional view of human-robot trust: our trust towards a robot may be influenced by the trust shown by the robot itself, following those reciprocal mechanisms that we generally observe in human-human interaction. In line with this view, social and collaborative robots should be able to adapt their trust-related behavior to modulate and maximize human trust. To achieve this goal, a social robot should be endowed with the ability to track human partners' capabilities and their trust in the robot itself. Moreover, it should react to the ongoing functional and relational joint dynamics to preserve or improve collaboration by increasing its trustworthiness in the eyes of the current human partner(s). Following reciprocal dynamics in interaction, the robot might need to balance the attempts to take the lead or comply with the human partner during a joint task, in order to optimize task-related performance and, at the same time, preserve human-robot trust-based collaboration and social norms.

In this respect, one important question is whether robots should exhibit *negative* reciprocity, that is, should distrust a human partner who does not trust the robot. In fact, this might appear as an anti-social behavior in the eyes of human partner: do we really want robots that distrust and possibly upset humans? The answer depends on the human-robot collaborative context and the relative goals of the interacting partners. In particular, we suggest that negative reciprocal trust could be useful in contexts in which robots must assist people in need (e.g., elderly people, patients with reduced mobility), who should trust the robot to accomplish their everyday goals. In human-human interaction, negative reciprocity has the peculiar function to signal to a selfish or anti-social partner the inappropriateness of their behavior. In many cases, the selfish individual re-starts to behave pro-socially to preserve the trust relationship. In the same way, a robot that negatively reciprocates trust (i.e., a robot that stop to trust the human partner when the human does not trust the robot) would signal that a social norm has been broken, possibly leading the human partner to increase their trust in the robot, with benefit for the human. Ironically, special attention should be put on the implementation of *positive* reciprocal trust in assistive robots. In this case, the robot would reciprocate trust by increasing its own level of trust in the human partner (e.g., a patient), possibly conceding more autonomy to the human. In this scenario, the robot should be careful in blindly reciprocating trust, since it should prioritize the patient's safety, even if this comes at the expenses of social norms.

In this regard, a key aspect concerning the development of "reciprocal" robots is the definition of the actual robots' goals, especially in the case of social robots that would collaborate with humans and assist them. Enabling robots to act with the unique goal of producing specific contextual actions (e.g., lifting a heavy object, accompanying a patient to a specific location) might lead back to an asymmetrical relationship between a human and a robot intended as a mechanical tool. To overcome this limitation, Man and Damasio [26] ambitiously suggested that robots, as intelligent and intentional agents, should hold their own meta-goal of self-preservation, acting as mechanical peers in human societies. This new class of autonomous agents would rely on homeostatic principles, which regulate body and mental states in order to maintain conditions compatible with life. At the same time, Man and Damasio argue that the goal of self-preservation should be combined with empathy, which would prevent robots to arm humans, or other robotic agents. We believe that further research should be conducted to investigate the impact of different robots' high-level goals on the human willingness to trust robots and collaborate with them. In this respect, a delicate issue is how a robot could manage a set of distinct, complementary goals in case of conflict between them. For instance, we need to understand how a robot should decide if reciprocating trust by balancing considerations on the immediate humans' emotional consequences of reciprocity with the long-term benefits of sustained human-robot collaboration.

Furthermore, it is still unclear whether humans should be aware of robots' goals and how this knowledge may influence the emergence of relational dynamics such as reciprocity. Can reciprocity arise when interacting with agents without transparent goals, motives and desires? In fact, reciprocity among humans settled as a social norm due to common knowledge on 1) the immediate, individual incentives to defect during cooperation and on 2) the complementary long-term benefits of cooperative behavior. On the contrary, knowledge of the motivations driving robots' actions can be extremely fuzzy in naïve individuals, possibly hindering the establishment of human-like relational mechanisms. For instance, reciprocity in human-robot interaction requires that humans know that robots are aware of social norms and may comply with them to achieve successful collaboration or please their human partner(s). In this sense, transparency about the purposes underlying the behavior of a robotic system may be crucial in promoting human-robot collaboration [27]. Further research is needed to understand whether a certain degree of transparency is necessary for the establishment of social norms in human-robot interactions. Human knowledge about the robot's goals and motives might be either general (e.g., the robot has the goal to preserve my safety and well-being) or domain-specific (e.g., the robot has the goal to help me get out of bed). Future studies may reveal the impact of partial or complete knowledge about different types of robots' goals on the establishment of social norms and the success of human-robot interaction and collaboration.

IV. CONCLUSION

The ambition of designing robotic collaborators, rather than anthropomorphic mechanical tools, opens the question of whether human-robot trust relationships should be reciprocal, as those among human peers. Although research on the role of reciprocity in human-robot trust is still very limited, recent findings suggest that trust towards robots is not a mere function of their perceived competence and reliability. Further research is needed to unveil the extent to which human trust in robots can be shaped by relational and reciprocal dynamics in joint activities. Crucially, these aspects may be fundamental in the design of robots that effectively act as collaborative companions in contexts such as healthcare, rehabilitation, education and assistance for the elderly.